\def\year{2021}\relax
\documentclass[letterpaper]{article} 
\usepackage{aaai21}  
\usepackage{times}  
\usepackage{helvet} 
\usepackage{courier}  
\usepackage[hyphens]{url}  
\usepackage{graphicx} 
\usepackage{natbib}  
\usepackage{caption} 

\usepackage{amssymb} 

\title{An Augmented Reality Platform for Introducing Reinforcement \\Learning to K-12 Students with Robots}
\author{
    Ziyi Zhang,
    Samuel Micah Akai-Nettey,
    Adonai Addo,\\
    Chris Rogers,
    Jivko Sinapov
    \\
}
\affiliations{
    \textsuperscript{\rm 1}Tufts University\\

    Medford, Massachusetts, 02155 \\
    \{ziyi.zhang|samuel.akai\_nettey|adonai.addo|chris.rogers|jivko.sinapov\}@tufts.edu

}

\begin{document}

\maketitle

\begin{abstract}
Interactive reinforcement learning, where humans actively assist during an agent's learning process, has the promise to alleviate the sample complexity challenges of practical algorithms. However, the inner workings and state of the robot are typically hidden from the teacher when humans provide feedback. To create a common ground between the human and the learning robot, in this paper, we propose an Augmented Reality (AR) system that reveals the hidden state of the learning to the human users. This paper describes our system's design and implementation and concludes with a discussion on two directions for future work which we are pursuing: 1) use of our system in AI education activities at the K-12 level; and 2) development of a framework for an AR-based human-in-the-loop reinforcement learning, where the human teacher can see sensory and cognitive representations of the robot overlaid in the real world.

\end{abstract}

\section{Introduction}
Artificial intelligence (AI) is playing a more prominent role in our daily life and has positively contributed to many fields and industries \cite{Fosel2018ReinforcementLW,doi:10.1021/acscentsci.7b00492}. Reinforcement Learning, in particular, has generated great interest over the past decade but still suffers from the challenge of sample complexity. To alleviate that issue, recent works have focused on how humans can assist robots with direct feedback (e.g., demonstrations or advice) during the learning process \cite{Ritschel2019AdaptiveLS,Elanie2021AFL}. Yet, the robot's inner workings and representations remain hidden from the users, which can result in a mismatch between the human's mental model of the robot's learning process and the actual inner workings on the robot. This paper proposes a tool based on Augmented Reality (AR) for the purpose of establishing common ground between a learning robot and its human teachers, with the goal of improving the efficacy of teaching and helping the human understand what the robot is learning. 

Our primary application is in AI education at the K-12 level. Reinforcement learning (RL) has contributed to many fields \cite{Bahare2018rlincontrol,Fosel2018ReinforcementLW} but has received little attention in K-12 AI education\cite{113eb79226ba49e987b923403eceacef}. However, basic RL concepts such as rewards for doing the right thing are close to our perception of human learning \cite{10.5555/3312046} and therefore may be easy for K-12 students to understand. Robots have been successfully used in K-12 STEM education.   \cite{Petrovi2021,Jeffery2013LEGOmiddleschool}. Some researchers have applied educational robots on teaching AI to undergraduate design students, improving students' engagement and lowering the barrier to entry in AI \cite{10.1007/978-3-540-69736-7_23}. One limitation of using physical robots is the lack of an intuitive way for users to visualize the AI training process and the robot's perception of the environment. We address this challenge by providing an AR tool that enables the user to visualize the robot's learning process and to actively assist in the training of the robot. 

Augmented reality (AR) is a technique to enhance the real world environment with virtual visual components, sounds or other sensory stimuli. By combining digital elements to a person's perception of the real world, AR is powerful on providing immersive experience and visualizing abstract concepts in physical world. Compared with virtual reality, AR does not require substantial hardware and can be easily implemented on different platforms that are easy to access including mobile phone, tablet, laptop, etc. In recent years, AR has been applied in various fields and industries including education \cite{Silva2019ARLA} and human robot interaction (HRI) \cite{Hennerley2017ARhri} which we would further introduce in the next section.

\begin{figure}[t]
    \centering
    \includegraphics[scale=0.095]{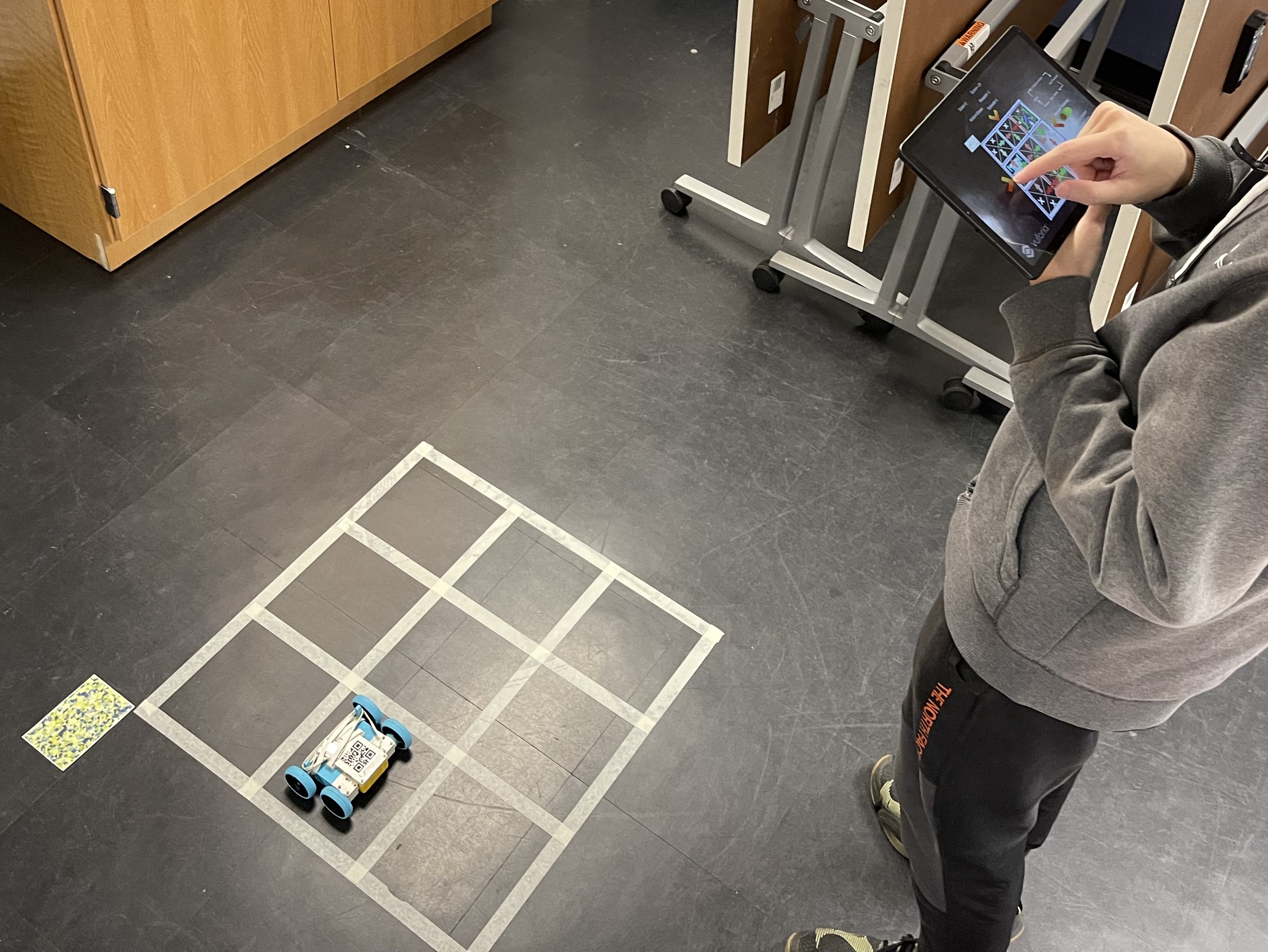}
    \caption{Users play with the RL activity and robot with the AR application installed in their mobile devices.}
\end{figure}

In this paper, we combined an AR interface with a LEGO\textregistered\ SPIKE Prime robot and designed a treasure hunting RL problem to introduce RL concepts to middle and high school students. The AR application can be easily installed on mobile devices (as showed in Figure 1) and the communication between the AR application and the robot is completed through an ESP8266 WiFi module connected to the robot. From the interface, the user can visualize internal RL representations and data structures. It also allows users to interpret key information related to training in an intuitive way. Since training a robot in physical world required longer time than training a virtual agent, we lowered the difficulty of our RL activity and gave users access to provide some instructions to the robot to decrease the sample complexity. The activity covers the following aspects of RL: 1) Key RL concepts, including state, action, reward; 2) Exploration and exploitation; 3) Q-table and how robots make decisions based on it; 4) Episodes and termination rules; and 5) Human-in-the-loop training. In the following sections we will unfold technical details and discuss potential applications of our system.

\section{Related Work}
In the past ten years, AR has gradually become a popular tool in K-12 education. Researchers have applied AR technique on teaching robotics \cite{cheli2018towards}, physics \cite{IBANEZ20141}, arts \cite{doi:10.1080/03004430.2015.1067888}, etc. And results from these studies showed that AR could not only increase students' engagement, concentration and learning outcome, but also reduce the difficulty of learning. However, currently AR is rarely used in K-12 AI education. Some researchers developed a virtual reality platform that could introduce RL to K-12 students in a digital world \cite{113eb79226ba49e987b923403eceacef}.

AR also contributed to many HRI research including robotic
teleoperation \cite{8594482}, robot debugging \cite{ikeda2021ar}, etc. Related work has proposed an AR visual tool named ``SENSAR" for human-robot collaboration \cite{DBLP:journals/corr/abs-2011-04515}, which was also deployed on a LEGO EV3 robot for educational purposes. The learning platform was an interface for students to visualize sensor values and control motors. In our research, we want to embed the learning experience into a game to make it more engaging, as other researchers also report greater engagement when the RL learning experience is centered around a game \cite{Matthew2011rlmario}.

\section{Reinforcement Learning background}

In Reinforcement Learning (RL), an agent has to
learn how to act based on scalar reward signals
detected over the course of interaction with its
environment. The agent's world is typically
represented as a Markov Decision Process (MDP), a
5-tuple $< \mathcal{S}, \mathcal{A}, \mathcal{T},
\mathcal{R, \gamma}>$, where $\mathcal{S}$ is a
set of states, $\mathcal{A}$ is a set of actions,
$\mathcal{T} : \mathcal{S} \times \mathcal{A}
\rightarrow \Pi(\mathcal{S})$ is a transition
function that maps the probability of moving to a
new state given action and current state,
$\mathcal{R} : \mathcal{S} \times \mathcal{A}
\rightarrow \mathbb{R}$ gives the reward of
taking an action in a given state, and $\gamma
\in [0,1)$ is the discount factor. We consider
episodic tasks in which the agent starts in an
initial state $s_0$ and upon reaching a terminal
state $s_{term}$, a new episode begins. 

At each interaction step, the agent observes its
state, and chooses an action according to its
policy $\pi : \mathcal{S} \rightarrow
\mathcal{A}$. The goal of the agent is to learn
an optimal policy $\pi^{*}$ that maximizes the
long-term expected sum of discounted rewards. One
way to learn the optimal policy is to learn the
optimal action-value function $Q^{*}(s, a)$,
which gives the expected sum of discounted
rewards for taking action $a$ in state $s$, and
following policy $\pi^{*}$ after:

$$Q^{*}(s,a) = \mathcal{R}(s,a) + \gamma \sum_{s'} \mathcal{T}(s'|s,a) \times max_{a'} Q^{*}(s',a')$$

Q-learning \cite{watkins1992q} is a common algorithm used to learn the optimal action-value function, where the Q-function is initialized arbitrarily (e.g., all zeros) and upon performing action $a$ in state $s$, observing reward $R$ and ending up in state $s'$, the Q-function is updated by:

$$Q(s,a) \leftarrow Q(s,a) + \alpha (R + \gamma \hspace{0.25em} max_{a'} Q(s',a') - Q(s,a))$$

\noindent where $\alpha$, the learning rate, is typically a small value (e.g., $0.05$). The agent decides which action to select using an $\epsilon$-greedy policy: with small probability $\epsilon$, the agent chooses a random action; otherwise, it chooses the action with the highest Q-value in its current state (i.e., the agent acts greedily with respect to its current action-value function). 

RL often requires large amounts of interaction with the environment. To speed up learning, researchers have proposed human-in-the-loop RL methods, e.g., learning-from-demonstration (LfD), where human teachers take over the action selection step, often providing several trajectories of complete solutions before the agent starts learning autonomously \cite{schaal1997learning}. In a related paradigm \cite{griffith2013policy}, the agent can seek ``advice'' from its human partner, e.g., what action to select in a particular state for which the Q-values are thought to be unreliable due to lack of experience. One of the goals of our system is to visualize the RL agent's knowledge and learning process as to provide information that makes human teachers more effective. 

\section{System Design}
Our system seeks to illustrate RL in an intuitive way as to provide students with a more immersive learning experience. There are three main components in our system, a LEGO SPIKE Prime robot, an AR application on a mobile device, and an ESP8266 WiFi module that is used for communication. We also designed a treasure hunting activity based on our platform in which students could train the robot to find treasure and escape through the exit.

\subsection{Hardware}
The educational robot, a LEGO SPIKE Prime, comes with different sensors (e.g., distance sensor and light sensor), and has 3 motors. The SPIKE Prime is customizable allowing students to modify it with additional sensors and pieces. However, one of the limitations is that it does not support WiFi. To solve this, we connected an ESP8266 microcontroller to one of the the robot's sensor port as an information transfer station. The ESP8266 board could communicate with the SPIKE through Universal Asynchronous Receiver-Transmitter (UART) protocol. Since the ESP8266 board has WiFi, we built an HTTP server on the board that could send and receive HTTP requests. Users could send robot data and commands by sending HTTP requests to the ESP8266 board and the board would transfer it to the robot through UART. On the other side, the robot could send its perception to the world to ESP8266 module through UART and the board would send a HTTP request that contains the data to the AR application for users to visualize. We have an image target at the corner of a physical grid for AR interface to track. Since the robot has to move a lot in the grid during training which can cause it to drift (usually 1 or 2 degrees after each move, also depends on the ground material),  we used the Inertial Measurement Unit (IMU) inside SPIKE to adjust its position and direction after each move.

\subsection{Activity Design}
The RL problem we designed is a grid world treasure hunting activity. The robot has to navigate itself in a 3$\times$3 grid to find the shortest path of getting the treasure and escape from the exit. In each state, the robot can only choose actions that won't make it across the boundary of the grid (e.g., robot can only move down or right when it is on the upper left corner). We have two virtual walls as showed in Figure 2 which robot can navigate to but unable pass. The robot is trained with a Q-learning algorithm that uses a $\epsilon$-greedy policy to make decisions. The reward system for robot finding the treasure, arriving the exit, hitting the wall or normal steps are +20, +30, -10, -1. Each episode ends when the robot successfully arrives at the exit no matter if it finds the treasure. While a 3x3 maze problem is a highly simplified version of an RL problem, we chose it because we wanted the user to be able to intuitively see the AI thinking that is happening in the AR experience.

The learning objectives of this activity were to let students: 1) Understand key RL concepts (state, action, reward and episode); 
2) Associate actions performed by the agent with the corresponding policy function, i.e., relate the robot's choice with values in the Q-table; 3) Understand the learning process of the robot in each training step; 4) Understand how humans can be part of the training process as to speed-up learning; and 5) Understand exploration and exploitation in the context of reinforcement learning.

\begin{figure}[t]
    \centering
    \includegraphics[scale=0.116]{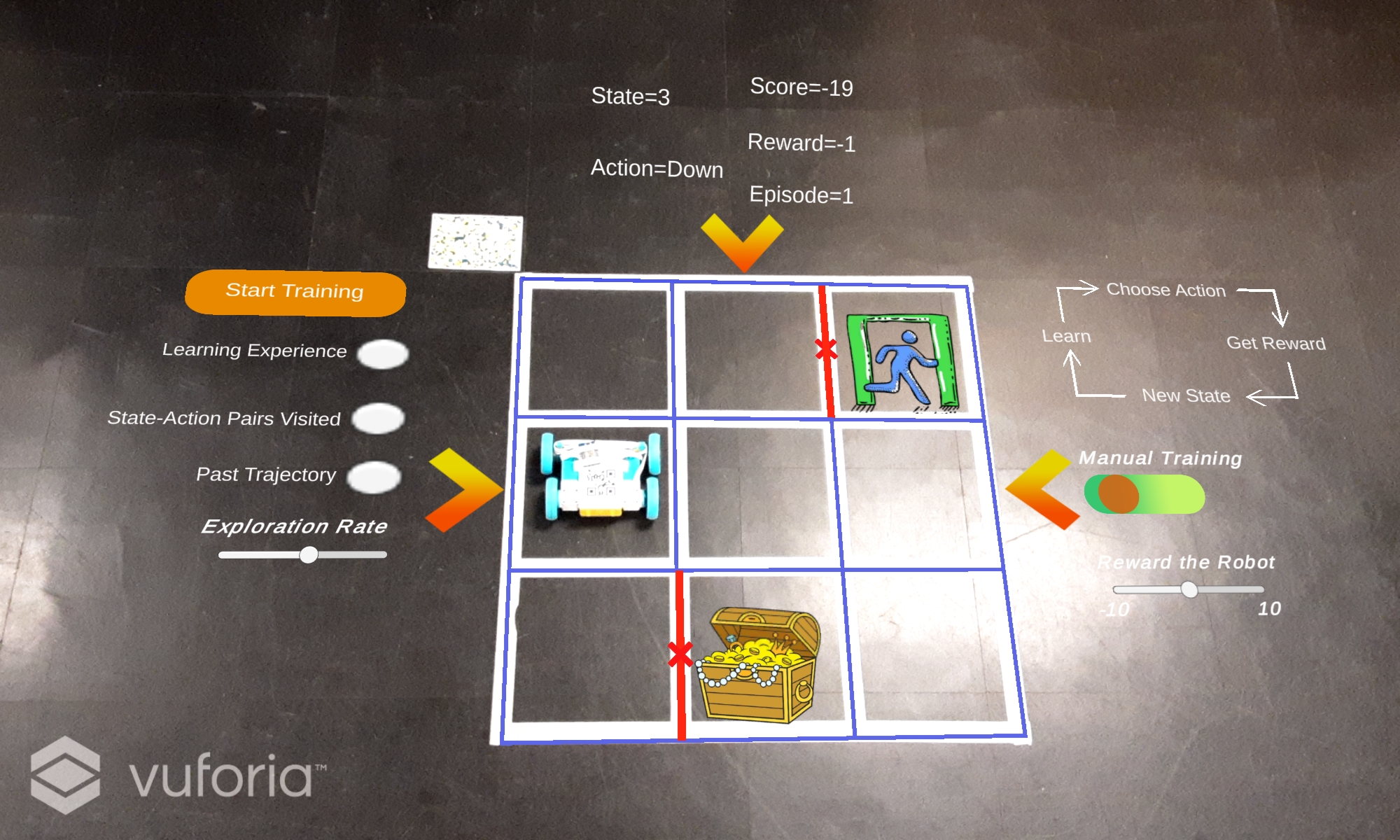}
    \caption{The interface was separated into three sections which students could use three buttons to fold and unfold each of them based on what they want to visualize.}
\end{figure}

\subsection{AR Interface}
Our AR application was developed using Unity\footnote{https://unity.com/} and Vuforia\footnote{https://developer.vuforia.com/}. The application can be deployed on iOS and Android mobile devices. Our interface contains four parts as shown in Figure 2; the left part contains a button for users to start training the robot and a slider to tweak the exploration rate ($\epsilon$). By default, the virtual grid in the middle shows the structure of the maze, but students could select what they want to visualize in the virtual grid through three checkboxes in the left UI. 
\begin{itemize}
    \item ``Learning Experience": If this checkbox is selected, a visualization of the Q-table will be displayed on the grid (Picture 1 in Figure 3). The arrows represents different possible actions in each state. As the training proceeds, the color of the arrows will update in real-time, e.g., a greener arrow represents a higher Q-value in the state-action pair, while a redder one represents a lower value.
    \item ``State-Action Pair Visited": This checkbox lets the virtual grid shows numbers of how many times each state-action pair has been visited by the robot (Picture 2 in Figure 3). This information can help students connect Q-values to training steps. If users want to train the robot with demonstrations, they can refer to this visualization to figure out in which states the robot needs more training.
    \item``Past Trajectory": By selecting this checkbox, an animation will show the trajectory the robot went through in the last episode (Picture 3 in Figure 3)
\end{itemize}
The upper part of the UI shows key RL parameters and variables used in training, including state, action, reward, episode and score (a gamification of accumulated reward that robot gets in current episode). 

The right portion of the UI contains the human-in-the-loop training interface, which allows users to switch between manual and automatic training modes. Most of the training would be done automatically, but we also want the users to provide instructions to the robot when it is making non-optimal actions during training process (e.g., unnecessary explorations, stuck in local optima). By switching to manual training mode, users can first give a reward to robot's last action using the slider, and this reward would replace the automatic reward received by the robot. Then users could tell the robot where it should go next by moving the robot in physical world and facing the robot to a desired direction. To detect the position and direction of the robot, we have another image target attached to the robot which could be tracked in the AR app. The manual training function allows users to participate in the training process through evaluating robot's past actions and guide the robot's future moves. This can make the activity more interactive and engaging, also help students understand how human input can be utilized in the training process. 

On top of the right UI is an animation to illustrate each training step. When in manual training mode, the current phase of the robot is highlighted both in the animation loop and the corresponding part in upper UI. For example, if the robot is waiting for the user to provide advice, the ``Choose Action" part in the loop and the ``Action" part in the upper UI will be highlighted. This part is for helping students obtain a deeper understanding on the RL training process.

Students can fold and unfold the upper, left and right part of the UI any time they want with three arrow buttons aside of the virtual grid so they won't get overwhelmed by all the information showing together on the UI. Considering the limitation of screen size, this also makes it more convenient for students to operate functions or observe information that they are interested in on mobile devices.

\begin{figure}[t]
    \centering
    \includegraphics[scale=0.47]{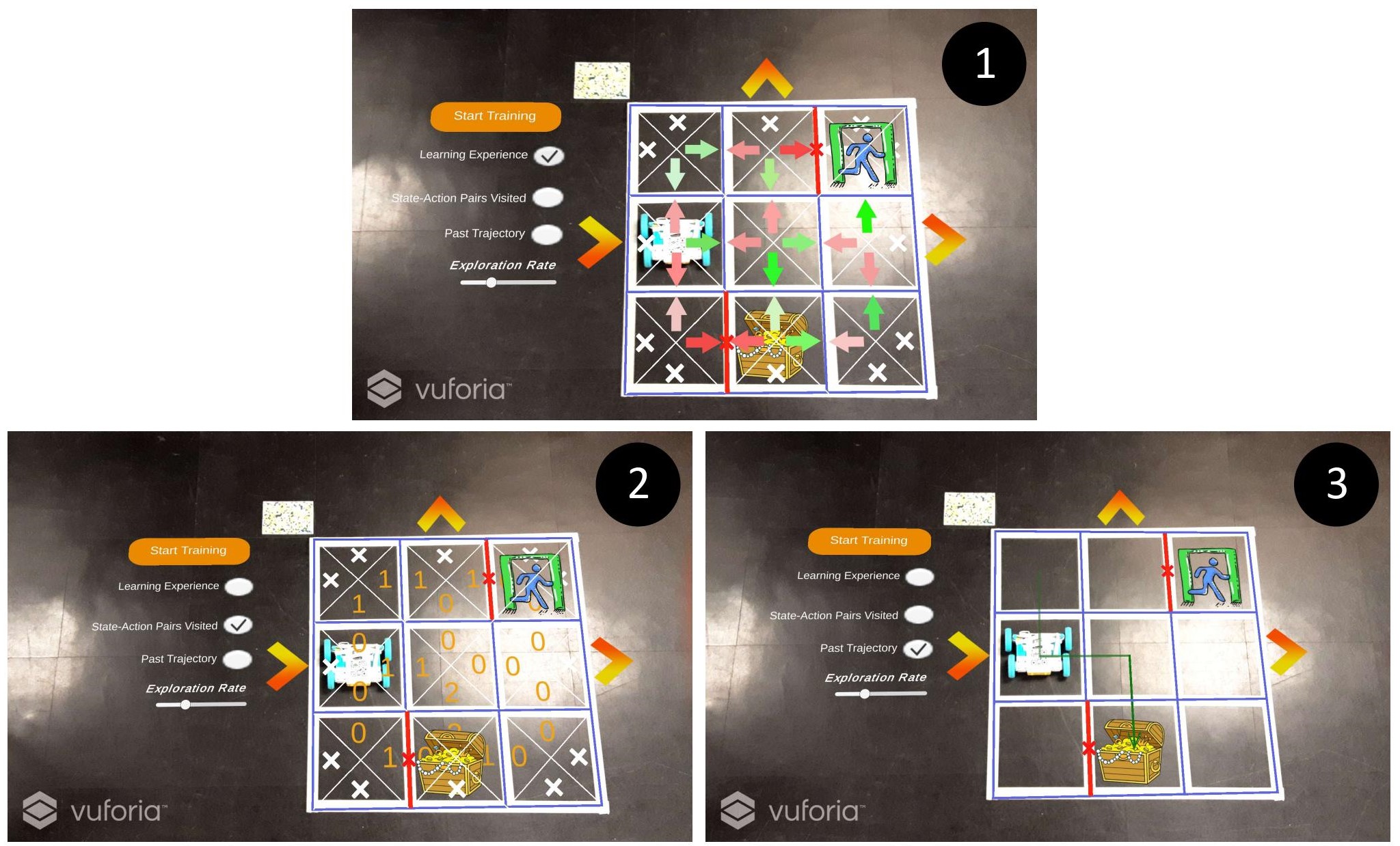}
    \caption{Different information that can be shown on the virtual grid. Including 1) a visualization of Q-table; 2) How many times each state-action pair has been visited; 3) An animation of Robot's trajectory in the last episode}
\end{figure}

\section{Proposed Applications}

\subsection{K-12 AI Education with Robots}
The main application of our system is AI education for K-12 students, specifically projects that leverage physical robots. The AR interface can be customized to demonstrate different AI concepts and provide visual aids for users to understand the robot's perception of the environment and the state of its learning process. Also, for some AI tasks that might take too long for the robot to perform all the training steps, the AR interface provides a way for users to do the training with virtual agents and implement the trained model on the robot to finish the task. Researchers and educators can use AR to design more engaging and immersive AI activities using physical objects to help students have a deeper understanding on how AI can be used to solve real world problems. Our platform also does not rely on desktop or laptop computers, which means students can try and learn AI conveniently on their mobile devices and in diverse environments.

\subsection{Human-in-the-loop Robot Training}
While training robots in the physical world with AI techniques, it is important to find an intuitive way for the robot to communicate its behaviors, plans and knowledge to humans, especially when humans need to assist in training the robot. Current platforms for human-in-the-loop RL typically hide the robot's representation from the human teacher. Our platform shows the possibility of using AR to build a straightforward and convenient HRI interface to help users finish human-in-the-loop robot training tasks. Compared with receiving information and controlling the robot with a traditional 2D screen, the AR interface can reduce unnecessary mental rotations from the screen to the physical world. The interface may also make it easier for users to track how the robot is interacting with the physical world and to provide appropriate feedback. We hypothesize that creating common ground with AR between a human teacher and a robot learner may result in more efficient teaching, thus reducing the number of interactions needed to learn a good policy.

\section{Conclusions and Future Work}
We proposed an AR robot platform for establishing common ground between a learning robot and a human user. We designed a treasure hunting activity based on it to introduce RL concepts to middle and high school students. The main significance of this project is to present an approach that can provide students a more interactive and immersive AI learning experience by combining physical robots with a virtual AR interface. We described the different components of our system and the RL activity we designed. We expect our platform can benefit other researches on K-12 AI education and provide some help on how to use AR in the visualization of human-in-the-loop AI training of robots. 

We are designing a pilot study with students this autumn to evaluate their learning outcome and get feedback to improve our platform. We plan to separate students into two groups, one group use our platform to learn AR and another group learn the same concepts through an online platform we designed before \cite{10.1007/978-3-030-82544-7_27}. The learning outcome of RL would be measured through a set of pre- and post- test. We would also compare the usability of two approaches and evaluate the value of AR and educational robots in K-12 RL learning. We are also working on constructing both easier and harder problems and create a storyline to let students explore them in a sequence. Finally, we also believe that visualizing the learning robot's internal representation may help human teachers be better at providing feedback to the robot and we plan to evaluate this hypothesis with an additional study. 

\section{Acknowledgments}
The research described in this paper was supported in part by a gift from the Verizon Foundation and LEGO Education.

\bibliography{references}

\end{document}